%% file: main.tex
\documentclass[11pt]{article}

\usepackage[margin=1in]{geometry}
\usepackage[T1]{fontenc}
\usepackage[utf8]{inputenc}
\usepackage{amsmath,amsthm,mathtools}
\usepackage{newtxtext,newtxmath}
\usepackage{microtype}
\usepackage{setspace}

\usepackage{booktabs,multirow,array,tabularx,makecell}
\usepackage{graphicx}
\usepackage{subcaption}
\usepackage{placeins}
\usepackage{xcolor}
\usepackage{enumitem}
\setlist{nosep,leftmargin=*}
\usepackage[font=small,labelfont=bf]{caption}

\usepackage[numbers,sort&compress]{natbib}
\usepackage{url}
\usepackage[colorlinks=true,linkcolor=blue!60!black,citecolor=blue!60!black,urlcolor=blue!60!black]{hyperref}
\usepackage[capitalize,noabbrev]{cleveref}

\usepackage{fancyhdr}
\theoremstyle{definition}

\theoremstyle{remark}

\input{math_commands}
\graphicspath{{figures/}}

\title{\LARGE\bfseries The Thinking Leakage in Reinforcement Learning for LLMs}

\author{%
Wenxuan Zhu$^{1}$\quad Shuming Liu$^{1}$\quad Bang An$^{1}$\quad Aznaur Aliev$^{1}$\\
Bing Li$^{1}$\quad Yibo Yang$^{2}$\quad Bernard Ghanem$^{1}$\\[0.6em]
$^{1}$King Abdullah University of Science and Technology\quad
$^{2}$Shanghai Jiao Tong University}
\date{}

\begin{document}
\maketitle

\begin{abstract}
\input{sections/0_abstract}
\end{abstract}

\input{sections/1_introduction}
\input{sections/2_related_work}
\input{sections/3_setup}
\input{sections/4_leakage}
\input{sections/5_attribution}
\input{sections/6_dynamics}

\input{sections/7_conclusion}

\clearpage
\bibliographystyle{plainnat}
\bibliography{references}

\clearpage
\appendix
\input{sections/A_appendix}

\end{document}

%% file: math_commands.tex
\newcommand{\R}{\mathbb{R}}

\newcommand{\think}{\textsc{think}}
\newcommand{\nothink}{\textsc{no-think}}
\newcommand{\pT}{p_{\mathrm{T}}}          
\newcommand{\pN}{p_{\mathrm{N}}}          
\newcommand{\base}{\pi_0}
\newcommand{\ck}[1]{\pi_{#1}}

\newcommand{\LLR}{\mathrm{LLR}}
\newcommand{\vth}{\mathbf{v}}             
\newcommand{\dk}{\boldsymbol{\delta}_k}            
\newcommand{\alphak}{\alpha_k}
\newcommand{\hres}{\mathbf{h}}
\newcommand{\Q}{\mathbf{Q}}
\newcommand{\mus}{\bar{\mathbf{h}}_l}

\newcommand{\dW}{\Delta W}

%% file: sections/0_abstract.tex
Hybrid reasoning models such as the Qwen3 series switch between a \think{} mode and a \nothink{} mode through a few chat-template tokens.
A popular recipe trains such models with reinforcement learning in \nothink{} mode and reports improved reasoning performance.
We provide evidence that much of this gain can be explained by \emph{thinking leakage}: the gain comes largely from partially reactivating the reasoning abilities of the original model's \think{} mode instead of learning enhanced reasoning behavior.
We train Qwen3-1.7B, 4B, and 8B with GRPO in \nothink{} mode. As accuracy improves, responses become longer and more reflective, are increasingly judged by the original model to resemble its own thinking, and the model's internal activations shift toward its \think{} mode.
We then investigate at the level of weights, activations, and attention heads: scaling the RL update further elicits \think{}-mode behavior and approaches \think{}-mode accuracy but does not exceed it; in Qwen3-8B, replacing a low-dimensional \think{}-related activation subspace of the RL-trained model with its original-model value removes most of the RL gain, and the reverse exchange transfers a considerable fraction of the gain to the original model; and the RL-induced shift passes through the attention heads that read the empty think block.
We further identify a possible mechanism in the GRPO training signal: responses favored by the advantage-weighted update are more \think{}-like, and a gradient step computed from this signal moves the model toward its \think{} state. Thus, gains in \nothink{} mode can largely reflect reactivation of reasoning abilities already present in the original model.
Code is available on \href{https://github.com/WenxuanZhu1103/think-leakage}{GitHub}.

%% file: sections/1_introduction.tex
\section{Introduction}
\label{sec:intro}

Reinforcement learning with verifiable rewards (RLVR) is widely used to improve the reasoning of large language models~\citep{deepseekr1,tulu3}. A rule-based verifier checks the model's final answer, and the model is trained on this reward, usually with Group Relative Policy Optimization (GRPO)~\citep{deepseekmath} or a variant of it~\citep{dapo,drgrpo}. 
This kind of training has produced large gains on reasoning tasks, together with long chains of thought~\citep{deepseekr1,longcot,prorl}.
Many recent reasoning models, however, are hybrid: the same model can work in either a \think{} or a \nothink{} mode. Examples include Qwen3~\citep{qwen3} and Llama-Nemotron~\citep{nemotron}; the Qwen3 series switches between a \think{} mode and a \nothink{} mode through a few chat-template tokens.
In \think{} mode the model first generates a long chain-of-thought between \texttt{<think>} and \texttt{</think>} before it answers. In \nothink{} mode the chat template itself appends an empty think block, \texttt{<think>\textbackslash n\textbackslash n</think>\textbackslash n\textbackslash n}, to the prompt, so that the model starts directly with its answer.

A common recipe applies RLVR to such a model in \nothink{} mode and reports the improvement in \nothink{} accuracy. But this improvement can come from different sources.
RL may teach the model new ways to solve problems, or, since the two modes share the same weights and differ only in the prompt, it may make the model use more of the reasoning behavior that already exists in its \think{} mode. We call the second effect \emph{thinking leakage}.
Prior work has asked whether RLVR adds reasoning ability beyond the base model~\citep{yue2025limit,oneexample,spurious,echo} and how a model can learn when to think~\citep{nothinking,thinkless,adaptthink}, but few studies have asked whether RL in one mode of a hybrid model draws on the other mode. In this paper, we show that the RL gain in \nothink{} mode comes largely from thinking leakage.

We train Qwen3-1.7B, 4B and 8B~\citep{qwen3} with GRPO on DAPO-Math-17k~\citep{dapo} in \nothink{} mode. As accuracy rises, each model moves toward its own \think{} mode. Its responses grow longer and contain more reflection words such as ``wait''. The original model increasingly judges the responses as \think{}-like; at 8B it prefers them to its own \nothink{} responses 94\% of the time.
The model's internal activations move in the same direction: at 8B, the shift reaches 0.56 of the activation difference between the original \nothink{} and \think{} modes.

However, these observations can not be directly attributed to the thinking leakage. We therefore design three experiments at the level of weights, activations, and attention heads. 
At the level of the weights, we scale the update that RL has learned and find that it amplifies \think{}-mode behavior and approaches, but does not exceed, the accuracy of the original model in \think{} mode. More interestingly, the scaled models write \texttt{</think>}, a token that nothing in \nothink{} mode calls for. 
At the level of the activations, we show that a low-dimensional subspace separating the two modes carries a large part of the RL gain; in Qwen3-8B, replacing this component with its original-model value removes 75\% of the improvement.
At the level of attention heads, we identify a small set of attention heads that keep the model in \nothink{} mode by reading the empty think block, and show that the RL-induced shift passes through these same heads.

Why, then, does RLVR drive the model toward thinking leakage?
We identify a possible source of this effect in the GRPO training signal: among responses to the same question, those favored by the advantage-weighted update are more \think{}-like than those it suppresses, and a gradient step computed from this signal moves the model toward its \think{} state. Thus, the optimization pressure toward thinking leakage is present in the training signal itself.

In summary, we study thinking leakage in RLVR for LLMs from three perspectives:

\begin{enumerate}
\item \textbf{Observed behavior.} We show that RL in \nothink{} mode moves hybrid reasoning models toward their existing \think{} mode, in both their generated responses and internal activations.
\item \textbf{Causal evidence.} Through interventions on weights, activations, and attention heads, we show that this shift largely comes from thinking leakage: the model reuses reasoning behavior from its \think{} mode, and this accounts for a large part of the RL gain in \nothink{} mode.
\item \textbf{Source of leakage.} We trace a possible origin of the effect to the GRPO training signal, which reinforces more \think{}-like responses more strongly and moves the weights toward the \think{} state.
\end{enumerate}

%% file: sections/2_related_work.tex
\section{Related work}
\label{sec:related}

\paragraph{RLVR for LLMs.}
Reinforcement learning with verifiable rewards trains LLMs using rewards generated by a rule-based verifier~\citep{tulu3,deepseekr1}, usually with a group-relative policy-gradient estimator such as GRPO~\citep{deepseekmath} or one of its variants~\citep{dapo,drgrpo}.
A growing body of work argues that much of what RLVR produces was already latent in the base model. \citet{yue2025limit} find that RL-trained models do not exceed their base model in pass@$k$.
A single training example~\citep{oneexample}, or even random and spurious rewards~\citep{spurious}, recovers most of the gain on Qwen models. \citet{echo} describe RL post-training as amplifying behaviors learned in pretraining, and \citet{negative} show that penalizing wrong answers alone is nearly as effective as the full signal. 

\paragraph{Hybrid reasoning, thinking budgets and latent reasoning.}
Models such as Qwen3~\citep{qwen3} support both thinking and non-thinking modes. \citet{nothinking} show that a reasoning model prompted with an empty thinking block remains surprisingly capable, which is the starting point of the present study: the \nothink{} template hides the capability rather than removing it. 
Other work learns when or how much a model should think. Some methods use RL to control the choice of reasoning mode~\citep{thinkless,adaptthink,lhrm}, while others control reasoning length~\citep{l1,s1,concise}. Test-time-compute scaling~\citep{ttc}, overthinking~\citep{overthink}, and long-chain-of-thought emergence under RL~\citep{longcot} study related trade-offs between reasoning length and performance. Reasoning can also be hidden from the visible output through latent tokens~\citep{quietstar} or continuous latent states~\citep{coconut}.

\paragraph{What RL changes in the weights.}
RL post-training changes the weights little and in a structured way: the update is sparse~\citep{subnet}, low-rank enough to be captured by small low-rank adapters (LoRA)~\citep{lora,lorawithoutregret}, avoids the principal directions of the pretrained weights~\citep{pathnottaken}, and forgets less than supervised fine-tuning because it stays close to the base policy~\citep{razor}. 
\citet{entropy} relate the collapse of policy entropy to the covariance between action probability and advantage.
Scaling a weight difference, $W_0 + c\,\Delta W$, is the basic operation of task arithmetic~\citep{taskarith}; we use it to ask where the RL update points. 
Related work argues that post-training largely changes how existing capabilities are elicited rather than adding new ones~\citep{lima,urial}; we ask whether this holds for the accuracy gain of RL in \nothink{} mode.

\paragraph{Directions, subspaces and interventions in the residual stream.}
A large body of work shows that model behavior can be represented and manipulated through directions in the residual stream. Representation engineering~\citep{repe} and activation addition~\citep{actadd,iti} use such directions to steer behavior. \citet{refusal} identify a direction that mediates refusal, while task and function vectors capture compact information about in-context tasks~\citep{taskvec,funcvec}.
Activation patching and causal mediation methods test whether particular activations are necessary for a behavior by replacing them with activations from another run~\citep{vig2020,rome,patching}. Attention knockout removes selected attention connections~\citep{knockout}, and causal-abstraction methods formalize what can be concluded from such interventions~\citep{causalabs,das}. Cross-model activation patching extends this idea to fine-tuned and base models~\citep{cmap}.
Our analysis follows this line of work. We define the think subspace using only the original model and training prompts. During free generation, we replace activations with values from the other model on the same tokens, so that position and content are matched. Related work has extracted reasoning steering vectors from distilled reasoning models~\citep{venhoff}, argued that base models already contain mechanisms later used by thinking models~\citep{basemodels}, and shown that reasoning fine-tuning can reuse existing representations~\citep{repurpose}. We go further by causally linking a think subspace and a small set of attention heads in the original model to the accuracy gains produced by RL in \nothink{} mode.

%% file: sections/3_setup.tex
\section{Experimental setup}
\label{sec:setup}

\paragraph{Models and templates.}
We experiment with Qwen3-1.7B, Qwen3-4B, and Qwen3-8B~\citep{qwen3}. We refer to these released checkpoints before our RL training as the \emph{original models}. In \think{} mode the prompt ends where the assistant's turn begins, and the model writes \texttt{<think>}, its reasoning, \texttt{</think>}, then the answer. In \nothink{} mode (\texttt{enable\_thinking=False}) the template appends \texttt{<think>\textbackslash n\textbackslash n</think>\textbackslash n\textbackslash n} to the prompt.

\paragraph{Training.}
At each step, we sample 256 prompts and generate 5 responses per prompt at temperature 1.
We use a PPO-style clipped objective~\citep{ppo} with clip ratio $0.2$, token-mean loss aggregation, and a KL penalty of $10^{-3}$ to the original model.
Each update uses four minibatches of 64 prompts. We train a rank-64 LoRA adapter~\citep{lora}, with scaling factor 64, on every linear layer. The optimizer is AdamW~\citep{adamw} with $\beta_1=0.9$, $\beta_2=0.999$ and weight decay $0.01$, a constant learning rate of $10^{-5}$ and gradient-norm clipping at $1.0$.

\paragraph{Evaluation.}
We evaluate on AIME 2022--2024, including both AIME I and II, for a total of 90 problems. Responses are sampled at temperature 1. We grade the last \texttt{\textbackslash boxed{}} expression against the reference answer; responses without a boxed answer are counted as incorrect.
All evaluations in both \think{} and \nothink{} modes allow up to 32{,}768 generated tokens per response. We refer to this as the 32k token budget.

\paragraph{Measures of think-likeness.}
Besides accuracy and response length, we use three measures of how much a response, or the model that produced it, resembles \think{} mode.
For a response $y$, let $\pT$ be the question rendered in \think{} mode followed by \texttt{<think>\textbackslash n}, and let $\pN$ be the same question rendered in \nothink{} mode, ending with the empty think block.
\begin{enumerate}
\item \textbf{Reflection words.} We count words and short phrases that indicate that the model is checking or revising its reasoning. These include \emph{wait}, \emph{hmm}, \emph{alternatively}, \emph{but wait}, \emph{actually}, sentence-initial \emph{no,}, and \emph{let me} followed by words such as \emph{check}, \emph{verify}, \emph{double-check}, \emph{think}, or \emph{reconsider}. Each position is counted at most once (\cref{app:measure}).
\item \textbf{Think win rate (TWR).} We use the original model $\base$ as a fixed judge. For a response $y$, we define
\begin{equation}
\LLR(y) = \frac{1}{|y|}\sum_{t=1}^{|y|}\bigl[\log \base(y_t \mid \pT, y_{<t}) - \log \base(y_t \mid \pN, y_{<t})\bigr],
\label{eq:llr}
\end{equation}
where $y_t$ is the $t$-th token of $y$, $y_{<t}$ the tokens before it, and the sum runs over all generated tokens of $y$, including the end-of-turn token when the response finished within the budget. 
A positive score means that the original model assigns higher per-token likelihood to $y$ under the \think{} prefix than under the \nothink{} prefix.
For a set of responses $S$, we define the \emph{think win rate} as
\begin{equation}
\mathrm{TWR}(S) = \Pr_{y \sim S,\; y' \sim \mathcal{R}_0}\bigl[\LLR(y) > \LLR(y')\bigr] + \tfrac{1}{2}\Pr_{y \sim S,\; y' \sim \mathcal{R}_0}\bigl[\LLR(y) = \LLR(y')\bigr],
\label{eq:twr}
\end{equation}
where $\mathcal{R}_0$ is the set of the original model's own \nothink{} responses.
This is the Mann--Whitney statistic applied to $\LLR$. The original model's own \nothink{} responses have a TWR of 50\% by construction. Its \think{}-mode text, measured using the first 4k tokens of each think block after removing the leading \texttt{<think>}, has a TWR of 100\%.
\item \textbf{Thinking activation shift $\alpha$.} For a layer $l$ (we report the middle layer, 18 of 36 for 8B and 4B, 14 of 28 for 1.7B) let $\mus(\pi, p, y)$ be the residual stream of model $\pi$ at layer $l$, that is, the vector of activations that layer $l$ passes on to the next layer, when the model reads $y$ after prefix $p$, averaged over all tokens of a fixed set of responses $y$, so that every token carries equal weight. The original model's \emph{think direction}, the difference between its activations in the two modes on the same text, and the \emph{shift} of the checkpoint after $k$ training steps are
\begin{equation}
\vth = \mus(\base,\pT,y) - \mus(\base,\pN,y), \qquad
\dk = \mus(\ck{k},\pN,y) - \mus(\base,\pN,y),
\end{equation}
both computed on the same fixed responses (the original model's own \nothink{} responses on the 90 AIME problems), so that the text is held constant token by token and only the prefix or the weights change. We summarize the shift by its coordinate along the think direction, which we call the \emph{thinking activation shift}
\begin{equation}
\alphak = \langle \dk, \vth\rangle / \|\vth\|^2 .
\label{eq:alpha}
\end{equation}
$\alphak=0$ means that the shift has no component along the think direction; $\alphak=1$ means that its component along that direction is as large as the full difference between the original model's \think{} and \nothink{} modes. $\alpha$ says nothing about the component orthogonal to $\vth$, which we also report ($0.58\,\|\vth\|$ at 8B, step 67). Because $\dk$ and $\vth$ share the term $\mus(\base,\pN,y)$, their inner product is biased upward when estimated on the same draws; every number is therefore also computed with cross-fitting (the direction is estimated on the even-numbered problems and the shift on the odd-numbered ones, and vice versa), which changes $\alpha$ by at most $0.01$.
\end{enumerate}

%% file: sections/4_leakage.tex
\section{RL in \nothink{} mode shifts the model toward \think{} mode}
\label{sec:leak}

\begin{figure}[!tbp]
\centering
\includegraphics[width=\linewidth]{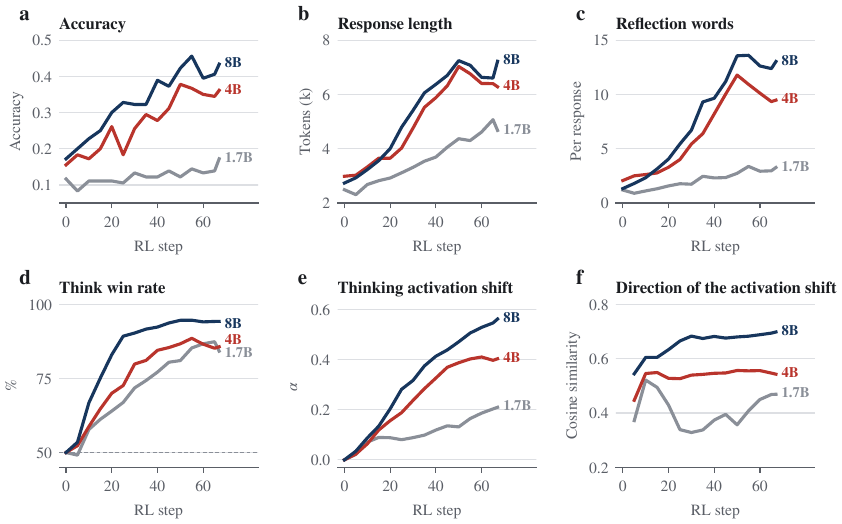}
\caption{\textbf{RL in \nothink{} mode shifts all three models toward \think{} mode.} Qwen3-8B, 4B, and 1.7B during training, evaluated in \nothink{} mode on AIME (90 problems, 2 samples per problem, 32k token budget). (a) Accuracy. (b) Mean response length. (c) Reflection words per response. (d) Think win rate (\cref{eq:twr}). (e) Thinking activation shift $\alpha$ (\cref{eq:alpha}) at the middle layer. (f) Cosine similarity between the activation shift and the think direction, on the same responses and layer as (e).}
\label{fig:traj}
\end{figure}

We first study how the three models change during RL training in \nothink{} mode. \Cref{fig:traj} tracks six quantities over training, all evaluated in \nothink{} mode.

\paragraph{Accuracy and response length increase together.}
Accuracy improves at all three model sizes, rising from 0.17 to 0.43 for Qwen3-8B (\cref{fig:traj}a), and responses grow longer at the same time, from 2.7k to 7.2k tokens for 8B, from 3.0k to 6.3k for 4B, and from 2.5k to 4.7k for 1.7B (\cref{fig:traj}b). They also contain more reflection words: the average count rises from 1.3 to 13.0 for 8B, from 2.1 to 9.5 for 4B, and from 1.2 to 3.3 for 1.7B (\cref{fig:traj}c). Both are characteristic of the original models in \think{} mode, whose responses under the same 32k budget contain 16k--18k tokens and about 120--130 reflection words on average, with accuracies of 0.74, 0.69, and 0.43 for 8B, 4B, and 1.7B.

\paragraph{The responses become more \think{}-like under the original model judge.}
The think win rate increases from 50\% to 94\% for 8B, 86\% for 4B, and 84\% for 1.7B (\cref{fig:traj}d), so after training an 8B checkpoint response receives a higher think-likeness score than an original \nothink{} response in 94\% of pairwise comparisons. The original model's own think-block text scores 100\%.

\paragraph{The internal activations also shift toward \think{} mode.}
The thinking activation shift $\alpha$ increases during training and reaches 0.56 for 8B, 0.40 for 4B, and 0.21 for 1.7B (\cref{fig:traj}e); for 8B, the activation change along the think direction is thus 56\% as large as the full difference between the original model's \nothink{} and \think{} modes, and the cross-fitted estimate is also 0.56. The same checkpoint has an orthogonal component of $0.58\,\|\vth\|$, so the full activation change is not limited to the think direction, and we therefore also measure the cosine similarity between the activation shift $\dk$ and the think direction $\vth$:
\begin{equation}
\cos(\dk,\vth) = \frac{\langle \dk,\vth\rangle}{\|\dk\|\,\|\vth\|}.
\end{equation}
A cosine of 1 means that the two directions are identical, while a value near 0 indicates little alignment. The cosine is positive at every checkpoint, ranging from 0.33 to 0.70 (\cref{fig:traj}f): for 8B it increases from 0.54 at step 5 to 0.70 at step 67, for 4B it stays near 0.55, and for 1.7B it ranges from 0.33 to 0.52. The RL-induced activation change is thus aligned with the original model's think direction at every checkpoint.

Across all three model sizes, the same pattern appears: as accuracy improves, responses become longer and more reflective, the original model judges them as more \think{}-like, and the internal activations move toward the original model's think direction. The effect is strongest for 8B, but is present at all three scales. These measures capture different aspects of the model, yet they change together during training. RL in \nothink{} mode therefore changes the model's output behavior and its internal state, and not only its accuracy, in the direction of \think{} mode.

%% file: sections/5_attribution.tex
\section{Evidence that the RL gain comes from thinking leakage}
\label{sec:attrib}

\Cref{sec:leak} shows that RL-trained models become more similar to their own \think{} mode. But this does not show that the accuracy gain comes from \think{} mode. RL could instead learn new problem-solving behavior that happens to look similar.
We test this with three interventions. First, we scale the weight update learned by RL and ask where it reaches (\cref{sec:update}). Second, we intervene directly on the activations that separate \think{} and \nothink{} modes (\cref{sec:causal}). Third, we identify the attention heads that attend to the empty think block and test whether the RL-induced change passes through the same heads (\cref{sec:circuit}).

\subsection{Weights: scaling the RL update brings the model close to \think{} mode}
\label{sec:update}

\begin{figure}[!tbp]
\centering
\includegraphics[width=\linewidth]{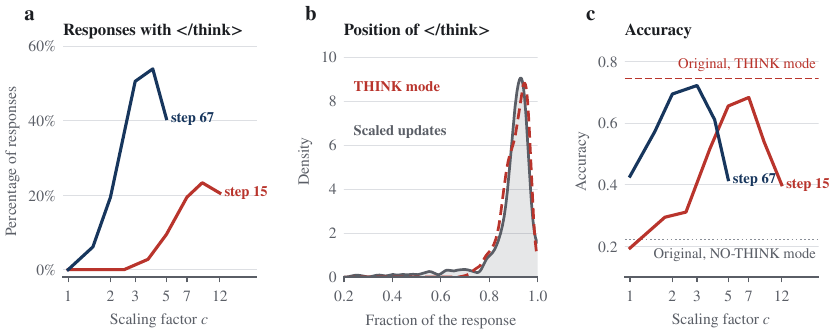}
\caption{\textbf{Scaling the RL update makes the model generate \texttt{</think>} and brings its accuracy close to \think{} mode.} Qwen3-8B with weights $W_0+c\,\dW$ from the step-15 and step-67 updates, evaluated on AIME in \nothink{} mode. (a) Fraction of responses containing \texttt{</think>}. (b) Position of \texttt{</think>} as a fraction of the response tokens (kernel density estimates), pooled over all scaled models (gray) and for the original model in \think{} mode (red, dashed). (c) Accuracy; the dashed line is the original model's \think{}-mode accuracy.}
\label{fig:amp}
\end{figure}

RL changes the model weights from $W_0$ to $W_0+\dW$, and if this update moves the model toward its existing \think{} mode, then scaling the same update should strengthen this effect. We therefore evaluate $W_0+c\,\dW$ for $c>1$, a standard operation on weight differences~\citep{taskarith}, without any further training. We use the Qwen3-8B updates after 15 and 67 training steps and evaluate all scaled models in \nothink{} mode with two samples per AIME problem (\cref{fig:amp}, \cref{app:amp}). This is an exploratory probe of the direction of the learned update: models at large $c$ are far outside the range seen during training, and the scaling factors were explored on the same 90 AIME problems.

\paragraph{The scaled models generate \texttt{</think>}, at the same place as in \think{} mode.}
In \think{} mode, the model generates \texttt{<think>}, its reasoning, \texttt{</think>}, and then the final answer. In \nothink{} mode the template has already closed the think block, the reward does not depend on this token, and the original model never generates it in 1{,}260 \nothink{} responses. The scaled models nevertheless begin to generate \texttt{</think>}: for the step-67 update it appears in 6\% of responses at $c=1.5$ and in 51\% at $c=3$, and for the step-15 update in up to 23\% (\cref{fig:amp}a). The token also appears where it appears in \think{} mode. Each of the 443 responses that contain it has it exactly once, near the end, at a median position of 91\% of the response against 92\% in \think{} mode, and the two distributions of this position almost coincide (\cref{fig:amp}b). No response contains \texttt{<think>}. As the RL update is scaled, the model thus closes a think block that was never opened, at the point where \think{} mode would close it (\cref{app:amp}).

\paragraph{Accuracy rises but does not exceed \think{}-mode accuracy.}
Accuracy also increases as the RL update is scaled, up to 0.72 for the step-67 update and 0.68 for the step-15 update, and at no tested value of $c$ does it exceed the original model's \think{}-mode accuracy of 0.74 (\cref{fig:amp}c).

\subsection{Activations: patching an 8-dimensional think subspace removes or transfers much of the gain}
\label{sec:causal}

We next test the RL-trained 8B model itself, asking whether its accuracy gain depends on the part of its activations that separates \think{} and \nothink{} modes.
We answer this by exchanging this part of the activations between the RL-trained model and the original model during generation.

\paragraph{The think subspace.}
We construct an 8-dimensional \emph{think subspace} at each layer using only the original model and training prompts. For 128 training prompts, we first generate a \nothink{} response with the original model. We then process the same response tokens once under the \think{} prefix and once under the \nothink{} prefix.
For layer $l$, the subspace $\Q_l\in\R^{d\times 8}$ is spanned by the mean activation difference between the two passes together with seven orthogonal directions that capture the largest remaining variation in the token-level differences (\cref{app:lesion}).
We also construct two 8-dimensional controls from the same tokens for comparison: a \emph{PCA} subspace, spanned by the eight directions of largest variance of the \nothink{} activations, and a \emph{random} subspace. 

\paragraph{Cross-model patching.}
Activation patching replaces activations of one run of a model with those of another run~\citep{vig2020,rome,patching}, and cross-model patching does so between a fine-tuned model and its original model on the same input~\citep{cmap}. We apply cross-model patching during free generation and restrict it to the subspace. One model is the \emph{target} and generates the response, while the other is the \emph{source} and processes exactly the same tokens. At the output of every layer $l=1,\ldots,35$ (all but the last of 36) and at every position, we replace only the target's component inside the think subspace:
\begin{equation}
\hres_{\text{target}} \leftarrow \hres_{\text{target}} + \Q_l \Q_l^{\!\top} \bigl(\hres_{\text{source}} - \hres_{\text{target}}\bigr).
\label{eq:edit}
\end{equation}
Everything outside the subspace is unchanged. In \emph{original-to-RL patching}, the RL-trained model generates the response but receives the original model's component inside the subspace. This removes the change made by RL within that subspace. In \emph{RL-to-original patching}, the original model generates the response but receives the RL-trained model's component. This transfers the RL-induced change to the original model. The subspace dimension and the patched layers were fixed before any AIME result was seen, and the pre-specified primary comparison was original-to-RL patching of the think subspace against original-to-RL patching of the PCA subspace.
Each condition uses two samples for each of the 90 problems with a 32k token budget, at which every response finishes.

\begin{figure}[!tbp]
\centering
\includegraphics[width=\linewidth]{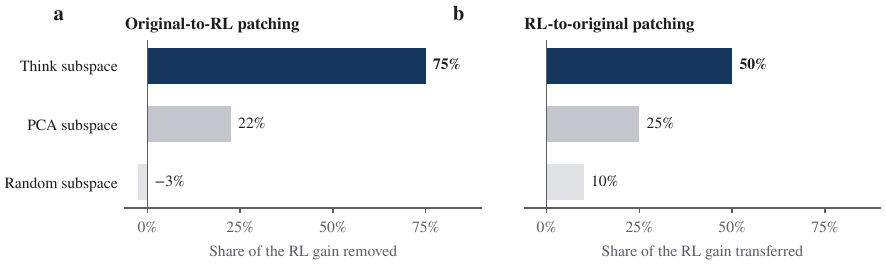}
\caption{\textbf{Exchanging the 8-dimensional think subspace removes or transfers a large part of the RL gain; PCA and random subspaces of the same size have much smaller effects.} Qwen3-8B at training step 67, evaluated on AIME in \nothink{} mode. (a) Original-to-RL patching: share of the RL gain $G$ removed. (b) RL-to-original patching: share of $G$ transferred.}
\label{fig:lesion}
\end{figure}

\paragraph{Results.}
\Cref{fig:lesion} summarizes the results. Let $G$ be the gain from RL, the accuracy of the unpatched RL-trained model minus that of the unpatched original model. The unmodified RL-trained model has an accuracy of 0.411 vs.\ 0.189 for the original model, giving $G=0.222$. Original-to-RL patching of the think subspace lowers accuracy to 0.244, removing 75\% of the RL gain, and response length falls from 7.0k to 4.2k tokens while reflection words fall from 13.0 to 3.9 per response. The reverse intervention transfers part of the gain: RL-to-original patching raises the original model's accuracy from 0.189 to 0.300, which is 50\% of $G$, with response length increasing from 2.7k to 4.5k tokens and reflection words from 1.6 to 5.3.

The control subspaces have much smaller effects: the PCA subspace removes 22\% of the gain and transfers 25\%; the random subspace has effects of $-3\%$ and 10\%.
In the pre-specified comparison, original-to-RL patching of the think subspace reduces accuracy by 0.117 more than original-to-RL patching of the PCA subspace. Patching only the single think direction removes 22\% of the gain, less than the 8-dimensional subspace. Increasing the subspace in RL-to-original patching from 8 to 16 dimensions raises the transferred share from 50\% to 60\%.

\subsection{Attention heads: the RL-induced shift passes through the heads that attend to the empty think block}
\label{sec:circuit}

The think subspace tells us where the mode difference appears in activation space; we next ask which parts of the network carry it. The only difference between the two prompts is the empty think block at the end of the \nothink{} template, the four tokens \texttt{<think>}, \texttt{\textbackslash n\textbackslash n}, \texttt{</think>} and \texttt{\textbackslash n\textbackslash n}. Information from these tokens reaches the response only through attention, so we first identify the heads that attend to the empty think block in the original model, and then test whether these same heads carry the change produced by RL.

We run both analyses for all three model sizes on the fixed responses of \cref{sec:setup}. The models read the first 2{,}048 tokens of each response without generating new tokens. Attention knockout uses all 90 responses. Single-head patching uses the first 30 responses to rank the heads and the remaining 60 as a held-out check. We write L$i$H$j$ for head $j$ of layer $i$, both counted from 1. Qwen3-8B and Qwen3-4B have 36 layers with 32 heads each (1{,}152 heads), and Qwen3-1.7B has 28 layers with 16 heads each (448 heads). Details are in \cref{app:heads}.

\paragraph{A small set of heads keeps the model in \nothink{} mode.}
For every head of the original model, we measure its \emph{attention to the empty think block}: the attention weight placed on \texttt{<think>}, \texttt{\textbackslash n\textbackslash n}, \texttt{</think>} and \texttt{\textbackslash n\textbackslash n}, summed over these tokens and averaged across response positions. This attention is highly concentrated. About 90\% of heads place less than 1\% of their attention on the block, while only five to seven heads in each model place at least 10\%.

To test whether this attention keeps the model in \nothink{} mode, we use attention knockout~\citep{knockout}. For a selected set of heads, we set the attention from response positions to the tokens of the empty think block to zero and renormalize attention over the remaining tokens. The prompt itself is unchanged. We measure two effects on the original model. The first is the thinking activation shift $\alpha$ (\cref{eq:alpha}). Without intervention, the original model has $\alpha=0$; its \think{} mode corresponds to $\alpha=1$. The second measures how much the intervention moves the model's next-token distribution toward that of \think{} mode, using the Kullback--Leibler (KL) divergence:
\begin{equation}
R_{\mathrm{KL}}(X) = 1 - \frac{\mathrm{KL}\bigl(p_{\mathrm{T}} \,\|\, p_X\bigr)}{\mathrm{KL}\bigl(p_{\mathrm{T}} \,\|\, p_{\mathrm{N}}\bigr)}.
\label{eq:klrec}
\end{equation}
Here, $p_{\mathrm{T}}$ and $p_{\mathrm{N}}$ are the original model's next-token distributions in \think{} and \nothink{} modes, and $p_X$ is its distribution in \nothink{} mode under intervention $X$. Each KL divergence is averaged over response positions from the second token on, since the first token is predicted at the last token of the block itself (\cref{app:heads}). Thus, $R_{\mathrm{KL}}=0\%$ means that the intervention leaves the \nothink{} distribution unchanged, while $R_{\mathrm{KL}}=100\%$ means that it matches the \think{}-mode distribution.

\begin{table}[!tbp]
\centering\small
\caption{\textbf{Removing attention to the empty think block moves the original model toward \think{} mode.} Thinking activation shift $\alpha$ (\cref{eq:alpha}) and relative KL reduction $R_{\mathrm{KL}}$ (\cref{eq:klrec}) of the original model under each intervention; random sets of heads are averaged over five draws.}
\label{tab:gate}
\input{tables/gate}
\end{table}

Knocking out attention to the empty block in all heads has almost the same effect as deleting the block itself (\cref{tab:gate}): $\alpha$ is 1.11, 1.17, and 1.13 for the 8B, 4B, and 1.7B models, and $R_{\mathrm{KL}}$ is 89\%, 85\%, and 90\%. Most of this effect comes from a small set of heads. Knocking out only the 5\% of heads that attend most strongly to the block gives $\alpha=0.78$, 0.88, and 0.66, and $R_{\mathrm{KL}}=72\%$, 74\%, and 77\% for the three model sizes. In contrast, knocking out the same number of randomly chosen heads gives an average $\alpha$ of at most 0.04 and an $R_{\mathrm{KL}}$ of at most 6\%. Thus, on these fixed responses, a small set of attention heads carries most of the signal that keeps the original model in \nothink{} mode.

\paragraph{RL changes the model through the same heads.}
We next test whether these heads also carry the change caused by RL, using cross-model patching of single heads. For each head, we let the original model process a fixed response in \nothink{} mode and replace that head's output with the output of the same head in the RL-trained model on the same tokens; if this single replacement moves the original model toward \think{} mode, its activation shift $\alpha$ becomes positive. We define the \emph{patching effect} of a head as
\begin{equation}
\frac{\alpha_{\text{original model with one head patched}}}{\alpha_{\text{RL-trained model}}},
\end{equation}
the fraction of the full RL-induced shift produced by patching that head alone. Effects from different heads are not additive. Because a head can only affect later layers, we measure $\alpha$ at three quarters of the network depth, layer 27 of 36 for the 8B and 4B models and layer 21 of 28 for the 1.7B model, and every head up to that layer is patched in turn.

\begin{table}[!tbp]
\centering\small
\caption{\textbf{The heads with the largest RL patching effects also attend strongly to the empty think block.} For each model size, the three heads with the largest patching effect, measured on the first 30 fixed responses; held-out values are in \cref{app:heads}. Attention to the empty think block: attention weight of the head on the tokens of the empty think block in the original model, averaged over response positions. Last column: rank of the head by this attention among all heads of the model.}
\label{tab:route}
\input{tables/route}
\end{table}

A single head can account for a large part of the activation shift (\cref{tab:route}): L18H19 has a patching effect of 44\% for 8B and 37\% for 4B, and L19H13 has an effect of 54\% for 1.7B. The same effects hold on the 60 responses that were not used to rank the heads: 44\%, 37\%, and 56\%, respectively, and the ordering of the top three heads is unchanged at every model size. These heads are among those that attend most strongly to the empty think block: L18H19 is in the top 0.2\% of heads by block attention in both the 8B and 4B models, and L19H13 is in the top 0.9\% for 1.7B. More generally, at every model size all three top patching heads belong to the top 5\% of heads by attention to the empty block, the set used in \cref{tab:gate}. Three heads chosen at random would contain 0.1 such heads on average (\cref{app:heads}).

The same heads that keep the original model in \nothink{} mode therefore also carry much of the RL-induced shift toward \think{} mode.

%% file: tables/gate.tex
\setlength{\tabcolsep}{4.5pt}
\begin{tabular}{@{}l cc cc cc@{}}
\toprule
& \multicolumn{2}{c}{Qwen3-8B} & \multicolumn{2}{c}{Qwen3-4B} & \multicolumn{2}{c}{Qwen3-1.7B} \\
\cmidrule(lr){2-3}\cmidrule(lr){4-5}\cmidrule(lr){6-7}
Intervention on the original model & $\alpha$ & $R_{\mathrm{KL}}$ & $\alpha$ & $R_{\mathrm{KL}}$ & $\alpha$ & $R_{\mathrm{KL}}$ \\
\midrule
Empty think block deleted from the prompt & 1.03 & 93\% & 1.11 & 88\% & 1.08 & 92\% \\
Attention to the block removed in all heads & 1.11 & 89\% & 1.17 & 85\% & 1.13 & 90\% \\
\addlinespace
Attention to the block removed in the 5\% of heads that attend most to it & 0.78 & 72\% & 0.88 & 74\% & 0.66 & 77\% \\
Attention to the block removed in 5\% of the heads, chosen at random & 0.03 & 3\% & 0.04 & 6\% & 0.02 & 3\% \\
\addlinespace
Attention to the block removed in the 1\% of heads that attend most to it & 0.59 & 53\% & 0.56 & 42\% & 0.11 & 25\% \\
Attention to the block removed in 1\% of the heads, chosen at random & 0.00 & 0\% & 0.00 & 1\% & 0.00 & 0\% \\
\bottomrule
\end{tabular}

%% file: tables/route.tex
\begin{tabular}{@{}llccc@{}}
\toprule
Model & Head & \makecell{Patching\\effect} & \makecell{Attention to the\\empty think block} & \makecell{Rank by attention\\to the block} \\
\midrule
\multirow{3}{*}{Qwen3-8B} & L18H19 & 44\% & 25.2\% & Top 0.2\% \\
 & L25H28 & 26\% & 8.6\% & Top 0.7\% \\
 & L20H5 & 17\% & 10.6\% & Top 0.6\% \\
\addlinespace
\multirow{3}{*}{Qwen3-4B} & L18H19 & 37\% & 27.5\% & Top 0.2\% \\
 & L20H5 & 17\% & 8.1\% & Top 1.0\% \\
 & L17H25 & 10\% & 3.6\% & Top 2.4\% \\
\addlinespace
\multirow{3}{*}{Qwen3-1.7B} & L19H13 & 54\% & 18.7\% & Top 0.9\% \\
 & L16H9 & 30\% & 8.7\% & Top 1.6\% \\
 & L20H15 & 24\% & 5.3\% & Top 2.7\% \\
\bottomrule
\end{tabular}

%% file: sections/6_dynamics.tex
\section{A possible cause: GRPO advantages favor \think{}-like responses}
\label{sec:mech}

The reward used in training depends only on whether the final answer is correct. \textbf{Why, then, does RL move the model toward \think{} mode?}
We study the advantages assigned during training, ask whether they favor more \think{}-like responses, and then take one gradient step using these advantages to measure whether it moves the model toward \think{} mode.

\paragraph{Data.}
We sample 20 responses for each of 256 training questions from the original model, using the same template and sampling settings as training. Each response receives reward $+1$ if the final answer is correct and $-1$ otherwise, and, as in training, responses are grouped in sets of five for the same question. The GRPO advantage of each response is its reward minus the group mean, divided by the standard deviation of the rewards in that group, so in a mixed group correct responses receive positive advantages and wrong responses receive negative advantages, with a magnitude that depends only on $k$, the number of correct responses among the five (\cref{tab:adv}). Groups in which all five responses have the same reward have zero advantage and are excluded.
We measure think-likeness using the original model as judge (\cref{eq:llr}).

\begin{table}[!tbp]
\centering\small
\caption{GRPO advantages for groups of five responses with reward $\pm1$, by the number $k$ of correct responses in the group.}
\label{tab:adv}
\input{tables/advantage}
\end{table}

\begin{figure}[!tbp]
\centering
\includegraphics[width=\linewidth]{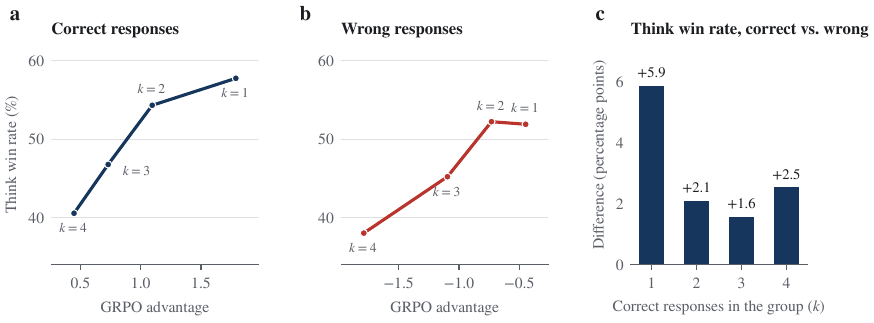}
\caption{\textbf{GRPO advantages favor more \think{}-like responses.} Qwen3-8B original model, 20 responses for each of 256 training questions, grouped in fives. (a, b) Think win rate of the correct (a) and wrong (b) responses against all sampled responses; each point is one value of $k$, placed at the advantage its responses receive. (c) Within groups with the same $k$, think win rate of the correct responses minus that of the wrong responses.}
\label{fig:mech}
\end{figure}

\paragraph{Among correct responses, the ones with larger advantages are more \think{}-like.}
A correct response in a group with only one correct answer receives $A=+1.79$, while a correct response in a group with four correct answers receives $A=+0.45$. The think win rate of correct responses decreases from 57.8\% at $k=1$ to 54.3\%, 46.8\%, and 40.5\% as $k$ increases to 4 (\cref{fig:mech}a), so correct responses that receive larger advantages are more \think{}-like. This is an association across groups rather than a direct effect of think-likeness on the advantage, which is determined only by $k$: groups with small $k$ tend to correspond to harder questions, and their correct responses are also more \think{}-like.

\paragraph{Among wrong responses, smaller penalties are associated with more \think{}-like responses.}
Wrong responses in groups with one correct answer receive the mildest penalty, $A=-0.45$, while the single wrong response in groups with four correct answers receives the strongest penalty, $A=-1.79$. The corresponding think win rates are 51.9\%, 52.2\%, 45.2\%, and 38.0\% as $k$ increases from 1 to 4 (\cref{fig:mech}b), so wrong responses that receive smaller penalties are also more \think{}-like. As above, this is an association across groups because the advantage is determined by $k$, not directly by the response's think-likeness.

\paragraph{Within the same group, reinforced responses are more \think{}-like than penalized responses.}
The within-group comparison is closer to the actual GRPO update because it compares responses to the same question. At every value of $k$, correct responses have higher think win rates than wrong responses, by 5.9, 2.1, 1.6, and 2.5 percentage points for $k=1,\ldots,4$ (\cref{fig:mech}c). Averaged over groups, the correct responses of a group rank 3.0 percentile points higher on the judge score than its wrong responses (\cref{app:dyn}). Because the advantages within each group sum to zero, each group's contribution to the gradient is a contrast between its correct and wrong responses, so differences between questions cancel within the group.

\paragraph{One gradient step moves the model toward \think{} mode.}
We next test where this training signal moves the model: we initialize a new LoRA adapter exactly as in training, compute the GRPO gradient on the responses, take one optimizer step at the training learning rate, and measure the change in the thinking activation shift $\alpha$ (\cref{eq:alpha}). This is a probe of the initial gradient, not an actual training step: it uses the responses collected for this analysis rather than a training batch. We use $\alpha$ because a single update changes generated text too little to measure reliably through sampling.

The full GRPO signal changes $\alpha$ by $+0.0075$ (\cref{tab:gradprobe}), positive advantages alone by $+0.0072$, and negative advantages alone by $+0.0057$. Thus, in this probe, both reinforcing correct responses and penalizing wrong responses move the model toward \think{} mode. We use two controls to test whether the effect depends on which responses receive which advantages. Shuffling the advantages within each group changes $\alpha$ by $-0.0022$, and giving every response the same weight $+1$ changes it by only $+0.0006$, so reinforcing all sampled responses equally produces little movement along the think direction. The shift appears to come mainly from the assignment of positive and negative weights to different responses.

\begin{table}[!tbp]
\centering\small
\caption{Change in the thinking activation shift $\alpha$ after one GRPO gradient step on a new LoRA adapter of the Qwen3-8B original model, under different weightings of the responses.}
\label{tab:gradprobe}
\input{tables/gradprobe}
\end{table}

%% file: tables/advantage.tex
\begin{tabular}{@{}lcccc@{}}
\toprule
$k$ = correct responses in the group of 5 & 1 (hard) & 2 & 3 & 4 (easy) \\
\midrule
Advantage of each correct response & $+1.79$ & $+1.10$ & $+0.73$ & $+0.45$ \\
Advantage of each wrong response & $-0.45$ & $-0.73$ & $-1.10$ & $-1.79$ \\
\bottomrule
\end{tabular}

%% file: tables/gradprobe.tex
\begin{tabular}{@{}lcc@{}}
\toprule
Weighting of the responses & Responses & $\Delta\alpha$ \\
\midrule
Full GRPO signal & 2{,}990 & $+0.0075$ \\
Positive advantages only (correct responses) & 1{,}479 & $+0.0072$ \\
Negative advantages only (wrong responses) & 1{,}511 & $+0.0057$ \\
\midrule
Advantages shuffled within each group & 2{,}990 & $-0.0022$ \\
Every response weighted $+1$ & 2{,}990 & $+0.0006$ \\
\bottomrule
\end{tabular}

%% file: sections/7_conclusion.tex
\section{Conclusion}
\label{sec:discussion}

In summary, our experimental results suggest that RL in \nothink{} mode does not simply improve a separate non-thinking policy. Instead, a large part of the improvement comes from thinking leakage.
This matters for how RL gains in hybrid reasoning models are interpreted. A higher \nothink{} accuracy does not necessarily mean that the model has learned better or new problem-solving strategies. It may instead reflect a change in mode control: behavior that was previously associated with \think{} mode becomes easier to express in \nothink{} mode.
The weight, activation, and attention-head interventions all support the above argument. They also suggest that the boundary between \think{} and \nothink{} is not fixed. RL can move this boundary even when training is performed entirely in one mode.
More broadly, these findings suggest that evaluations of reasoning post-training should separate two questions: whether RL adds new capability, and whether it changes how existing capability is accessed. For hybrid reasoning models, this distinction is especially important because the two behaviors are implemented by the same network and can be only a few prompt tokens apart.

%% file: sections/A_appendix.tex
\section{Measurement details}
\label{app:measure}

\paragraph{Prefixes and token sequences.}
Both modes use the Qwen3 chat template. $\pN$ is the question rendered with \texttt{enable\_thinking=False}; it ends with the assistant turn marker followed by the empty think block quoted in \cref{sec:setup}. $\pT$ is the question rendered with thinking enabled, which ends with the assistant turn marker, followed by \texttt{<think>} and a newline, so that a response placed after it is scored as the content of a think block. All measures are computed from the generated token ids, including the end-of-turn token when the response finished within the budget.

\paragraph{Reflection words.}
We use case-insensitive regular expressions over the response text that match whole words: \emph{wait}; \emph{hmm} (any number of m's); \emph{alternatively}; \emph{let me} followed by any \emph{re-} word or by \emph{check}, \emph{verify}, \emph{double-check}, \emph{think} or \emph{reconsider}; \emph{but wait}; \emph{actually}; and a sentence-initial \emph{no,}. The count is the number of non-overlapping matches of the alternation of these patterns, so a position matched by two patterns (\emph{but wait} and \emph{wait}) is counted once. Counts are reported per response.

\paragraph{Think win rate.}
The judge score of \cref{eq:llr} is computed with the original model by teacher forcing over the full response: the model reads the prefix followed by the response and assigns a log-probability to every response token. The two prefixes differ only in the tokens that select the mode. The win rate of a set of responses is the Mann--Whitney statistic of its per-response scores against those of the original model's \nothink{} responses, pooled over the 90 problems (all samples on both sides; ties count half). The variant used in \cref{sec:mech} takes the pool of all responses sampled for the training questions as the reference. The judge separates the original model's own \think{}-mode text from its \nothink{} responses completely (win rate 100\%; the scores differ by about 0.08 nats per token at 8B). The \think{}-mode text used for this reference point is the content of the original model's think block, generated with a 4k budget and with the leading \texttt{<think>} stripped. Most of these samples are truncated before \texttt{</think>}, so the reference contains reasoning text only, without \texttt{</think>} or the final answer.

\paragraph{Thinking activation shift.}
Activations are read from the output of each layer (the \texttt{hidden\_states} of the Hugging Face implementation; index $l$ is the output of layer $l$, counting layers from 1). $\mus(\pi, p, y)$ is the mean over all response tokens of the fixed responses pooled together, so that every token carries equal weight. The fixed responses are the original model's own \nothink{} responses on the 90 problems, the first sample per problem, and $\vth$ and $\dk$ are computed on the same token ids. Cross-fitting splits the fixed responses by the parity of the problem index. At 8B, step 67, the norm of the part of the shift that is orthogonal to $\vth$ is $0.58\,\|\vth\|$. We report layer 18 of 36 (14 of 28 for 1.7B). At layer 9 the coordinate is close to zero at 8B and small at 4B, as expected from the think component appearing only from layers 12--14 onward (\cref{app:write}); at layer 27 it rises along training as at layer 18, and is larger than at the middle layer for 8B and smaller for 4B.

\FloatBarrier

\section{Scaling the RL update}
\label{app:amp}

\Cref{tab:amp} lists all scaled models of \cref{fig:amp}. The step-15 update at $c=7$ and the step-67 update at $c=2$ reach the same thinking activation shift ($\alpha = 0.96$ and $0.98$) and give the same accuracy (0.68 and 0.69) and the same share of responses with \texttt{</think>} (19\% and 19\%), although the two scaled updates differ in size, measured by the Frobenius norm of the weight change, the square root of the sum of its squared entries ($7 \times 0.51 = 3.6$ against $2 \times 0.97 = 1.9$). This matched pair suggests that the position of the activations along the think direction is more informative than the size of the weight update, although two points cannot establish this in general. The scaling factors were explored on the same 90 problems, so the peaks at $c=7$ (step 15) and $c=3$ (step 67) are selected values and not held-out ones.

\begin{table}[h]
\centering\small
\caption{Qwen3-8B with the step-15 and step-67 updates scaled by $c$. AIME, \nothink{} mode, 32k budget, two samples per problem. Finished: share of responses that end within the budget.}
\label{tab:amp}
\setlength{\tabcolsep}{4pt}
\input{tables/amp}
\end{table}

\paragraph{How the scaled models use \texttt{</think>}.}
Over the 14 scaled models (2{,}520 responses), 443 responses contain \texttt{</think>} and none contains \texttt{<think>}. Each of the 443 contains the token exactly once. Its position, as a fraction of the tokens of the response, has a median of 0.91 (interquartile range 0.88--0.94); in the original model's \think{} mode the median is 0.92 (0.88--0.95; 667 responses that contain the token). \Cref{fig:amp}b shows the two distributions. The text after the token has a median length of 1.8k tokens (1.1k in \think{} mode) and most often begins with the words ``To solve the problem, we are given'', which are also the most frequent beginning of the final answer in \think{} mode. For comparison, the original model writes \texttt{</think>} in none of its 1{,}260 \nothink{} responses (four independently generated sets of responses to the 90 problems, three with two responses per problem and one with eight).

\begin{figure}[h]
\centering
\includegraphics[width=0.72\linewidth]{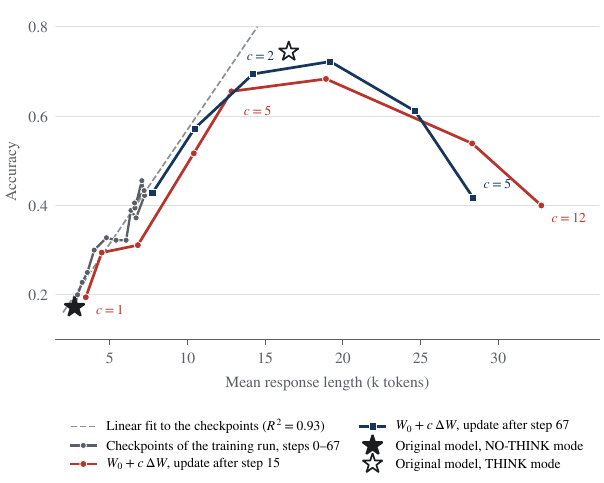}
\caption{Accuracy against mean response length, Qwen3-8B on AIME with a 32k budget. Gray: training checkpoints in \nothink{} mode with a linear fit (dashed). Red and navy: the step-15 and step-67 updates scaled by $c$. Stars: the original model in \nothink{} mode (filled) and \think{} mode (open; eight samples per problem, the other points use two). The three sets differ in mode and sample count, so only their shapes are comparable.}
\label{fig:lenlaw}
\end{figure}

\paragraph{Accuracy and response length.}
\Cref{fig:lenlaw} places the scaled models next to the checkpoints of the training run. Up to $c=5$ for the step-15 update and $c=2$ for the step-67 update, the scaled models follow, slightly below it, the relation between accuracy and mean response length that the checkpoints themselves trace: scaling continues the change that training makes. Beyond these factors the responses reach the 32k budget, the share that finishes falls below 0.9, and accuracy falls far below that relation.

\FloatBarrier

\section{Think subspace and cross-model patching}
\label{app:lesion}

\paragraph{Estimation of the subspaces.}
The subspaces are estimated on 128 DAPO-Math-17k prompts (drawn at random, seed 0) with one \nothink{} response of the original model each, using the first 512 response tokens. The same token ids are processed after (T) the \think{} prefix and (N) the \nothink{} prefix. For each layer, the second moment of the per-token differences $\hres_{\mathrm T} - \hres_{\mathrm N}$ is accumulated after dropping the first four response tokens and normalizing each difference vector to unit length. Both steps are needed. A handful of tokens have very large activations, and without these steps the leading eigenvector is a single outlier token (it accounts for 0.998 of the second moment and has a cosine of 0.31 with the mean difference). The first basis vector of the think subspace is the normalized mean of the raw differences, and the other seven are the leading eigenvectors of the second moment after that direction has been projected out. The PCA subspace consists of the top eight eigenvectors of the centered covariance of the unit-normalized $\hres_{\mathrm N}$, with the same treatment of tokens; it is a control for the amount of variance and is not matched to the think subspace in any other quantity. The random subspace is spanned by a random orthonormal $d\times 8$ matrix per layer, drawn uniformly among all 8-dimensional subspaces (seed 1234). The overlap between the think and the PCA subspace, measured as the mean squared projection of one orthonormal basis onto the span of the other, is 0.08--0.22 across layers.

\paragraph{Patching.}
\Cref{eq:edit} is applied to the output of layers $1, \dots, 35$ (the output of the last layer only feeds the final normalization and is not patched), at every position, including the positions of the prompt, and to the target model only. The source processes the same tokens and is never patched. The component inside the subspace is replaced completely. Both models keep their own key--value caches during generation, so the cache of the target contains patched activations. Responses are sampled at temperature 1, two per problem, first with an 8k budget. Every response that reached 8k tokens was then continued under the same patch up to 32k: the stored text is tokenized again (the result is identical to the original token ids in 171 of 180 checked responses and differs by one merged token in the others), the prompt and the stored response are processed again by the source and by the patched target, and generation resumes with a new random seed. Because the patched activations are a deterministic function of the tokens, this reproduces the original computation up to the nine re-tokenized responses, and the responses that finished within 8k together with the continued ones form a sample of the patched model at the 32k budget. Sampling the truncated responses again from scratch would instead over-represent short responses. At 32k every response of every condition finishes. RL-to-original patching of a 16-dimensional think subspace (the same basis extended by the next eight eigenvectors) was run in the same way and gives an accuracy of 0.322 (length 4.7k tokens, every response finished), an effect of $+0.133$ or 60\% of $G$. The dimension of the subspace, the set of patched layers and the complete replacement were fixed before any AIME result was seen. The two unpatched models were sampled anew for this experiment, so their accuracies (0.411 and 0.189) come from a sample that is independent of the one in \cref{sec:leak} (0.433 and 0.172); the differences between the two samples are within sampling variation. \Cref{tab:lesion2} lists every condition.

\begin{table}[h]
\centering\small
\caption{All conditions of the cross-model patching experiment. Qwen3-8B, training step 67, AIME, 32k budget, two samples per problem. Share of $G$: share of the RL gain $G=0.222$ removed (original-to-RL) or transferred (RL-to-original). Finished: share of responses that end within the budget. Subspaces have 8 dimensions per layer unless stated.}
\label{tab:lesion2}
\setlength{\tabcolsep}{4pt}
\resizebox{\linewidth}{!}{\input{tables/lesion_seed2_32k}}
\end{table}

\FloatBarrier

\section{Steering the original model}
\label{app:steer}

In this experiment a multiple $a\,\vth^{\text{own}}$ of a think direction is added to the output of layer 18 of the 8B original model at every generated position, in \nothink{} mode. Here $\vth^{\text{own}}$ is estimated from the original model's own responses in the two modes (the mean activation over its \think{}-mode responses minus the mean activation over its \nothink{} responses) and not on fixed responses as in \cref{sec:setup}. One response per problem was generated with a 4k budget, and every response that reached 4k tokens was then continued to 32k under the same steering (the stored text is tokenized again and processed with the vector added at the positions of the response). \Cref{tab:steer} gives the results. The number of reflection words, the response length and the accuracy all rise with $a$. Accuracy goes from 0.19 at $a=0$ to 0.60 at $a=1$, where 24\% of the responses reach the budget. At $a=0.5$ the number of reflection words per response (12.4) is that of the RL-trained model (12.4--13.6 at steps 50--67), while accuracy is 0.29 against 0.43.

\begin{table}[h]
\centering\small
\caption{Steering the 8B original model along its think direction at layer 18. AIME, \nothink{} mode, 32k budget, one sample per problem. Finished: share of responses that end within the budget.}
\label{tab:steer}
\input{tables/steer}
\end{table}

\FloatBarrier

\section{Attention-head experiments}
\label{app:heads}

\paragraph{Texts.}
Both analyses of \cref{sec:circuit} use the fixed responses of \cref{sec:setup}, truncated to their first 2{,}048 tokens. The models read these recorded tokens and do not generate (teacher forcing). Because of the truncation, the values of $\alpha$ differ slightly from those of \cref{sec:leak} (0.52 instead of 0.56 for the RL-trained 8B model at the middle layer).

\paragraph{Attention to the empty think block.}
For each head of the original model in \nothink{} mode, the attention weights from a response position to the tokens of the empty think block are summed, averaged over the response positions of a text and then over the 90 texts. The attention is concentrated in few heads: 91\%, 91\% and 90\% of the heads (8B, 4B, 1.7B) place less than 1\% of their attention on the block, and only 7, 7 and 5 heads place 10\% or more. The three largest values at 8B are 64\% (L4H23), 25\% (L18H19) and 18\% (L12H4).

\paragraph{Attention knockout.}
In the chosen heads, the attention scores from every response position to the positions of the empty think block are set to the most negative representable value before the softmax, so that the attention weights on the block become zero and the remaining weights are renormalized. Nothing else is changed. The knockout acts on the response positions only. The first response token is predicted at the last token of the block itself, which the knockout does not affect, so $R_{\mathrm{KL}}$ is computed over the predictions of the second response token onward. The condition in which the block is deleted removes its tokens from the prompt and is evaluated with the same metric. The heads of the ``5\%'' and ``1\%'' conditions are the heads with the largest attention to the block among all heads of the model, a criterion that does not involve the outcome of the knockout. Random sets of heads are drawn uniformly among all heads, and we report the mean over five draws. Among the five draws of 5\% of the heads, the largest $\alpha$ is 0.14, 0.18 and 0.11 (8B, 4B, 1.7B) and the largest $R_{\mathrm{KL}}$ is 10\%, 10\% and 12\%. At the first response token, which is excluded, the KL divergence between the two modes is 17 to 20 nats, against 0.02 to 0.03 nats per token over the following tokens.

\paragraph{Head patching.}
The original model reads the fixed responses in \nothink{} mode. For one head at a time, the output of the head (its part of the input of the layer's attention output projection) is replaced at all positions by the output of the same head in the RL-trained model (training step 67) on the same tokens. The thinking activation shift of the patched model is measured at layer 27 of 36 (8B, 4B) or 21 of 28 (1.7B), three quarters of the depth, because a head above the middle layer cannot change the activations at the middle layer. For the same reason only the heads up to the measurement layer are patched: 864 heads at 8B and 4B and 336 at 1.7B. The patching effect of a head is $\alpha$ of the patched original model divided by $\alpha$ of the RL-trained model, both computed on the first 30 of the 90 texts. The heads are thus ranked and their effects are measured on the same 30 texts. As a check, we measured the ten heads with the largest effects of each model again on the other 60 texts, which played no part in the ranking. The effects agree. For the five heads with the largest effects they differ by at most 0.6, 1.0 and 1.5 percentage points (8B, 4B, 1.7B), the order of the five heads is the same, and the largest effects are 44\%, 37\% and 56\% (against 44\%, 37\% and 54\% on the 30 texts). \Cref{tab:route} lists the three heads with the largest patching effects at each model size. The effects of single heads are not additive (their sum over all patched heads is 2.5, 1.7 and $-0.4$ at 8B, 4B and 1.7B), and some heads have negative effects; the most negative are $-8\%$ (L17H28, 8B), $-14\%$ (L23H15, 4B) and $-35\%$ (L20H16, 1.7B). The 5\% of all heads that attend most to the empty think block are 58 heads at 8B and 4B and 22 heads at 1.7B, of which 34, 38 and 14 lie at or below the measurement layer and are therefore among the patched heads. Of the three heads with the largest patching effects, all three belong to this set at every model size; of the five, 5, 4 and 5; and of the ten, 9, 8 and 5. If the three heads were drawn at random among the patched heads, the expected number in the set would be 0.1.

\FloatBarrier

\section{Where the activation shift is written}
\label{app:write}
Only two kinds of modules write into the residual stream: the output projections of the attention layers and the output (down) projections of the MLPs. For each such module, the output of the RL-trained model on a token is $W_0 x' + \Delta W x'$, where $x'$ is its input, which differs from the input $x$ in the original model. The change of the output relative to the original model is therefore $(W_0 x' - W_0 x) + \Delta W x'$: a term produced by the frozen pretrained weights acting on a changed input, and a term produced directly by the update. We accumulate both terms along the think direction $\vth$ and orthogonal to it over the fixed responses; their sum agrees with the measured shift to within 2\% (median over layers). More than 99\% of the shift along $\vth$, and of the orthogonal shift, comes from the frozen-weight term. The component along $\vth$ is zero in the first ten layers and appears at layers 12--14. Consistently, removing from the update of these modules the part that writes into the think subspace changes nothing (\cref{tab:surgery}): that part contains 0.6\% of the squared norm of the update, no more than a random subspace. Removing the top eight singular directions of the update, or keeping the update of only the modules that write into the residual stream, does lower accuracy. The dominance of the frozen-weight term is a general property of a small update whose effect is propagated through a large network. We report it as a statement about where the shift arises and not as evidence for thinking leakage: it does not by itself distinguish the re-use of pretrained computation from new computation.

\begin{table}[h]
\centering\small
\caption{Removing parts of the RL update of Qwen3-8B (training step 67). AIME, \nothink{} mode, 32k budget, eight samples per problem. The part that writes into a subspace is the output component of the update in that subspace (8 dimensions per layer). q, k, v, o: attention projections; gate, up, down: MLP projections.}
\label{tab:surgery}
\input{tables/surgery}
\end{table}

\FloatBarrier

\section{Training-signal analysis}
\label{app:dyn}

\paragraph{Sampling and scoring.}
256 training questions are drawn once (seed 1). The 8B original model answers each question 20 times with the training template, temperature and length limit. Each response receives the training reward $r \in \{-1, +1\}$ (the DAPO scorer, which requires the final answer on a line that begins with \texttt{Answer:}). The responses are divided into groups of five by sample index (four groups per question) and receive the GRPO advantage as implemented in verl, $A_i = (r_i - \bar r_g)/(\mathrm{sd}_g + 10^{-6})$, with the unbiased standard deviation $\mathrm{sd}_g$ of the rewards in the group. Groups in which all rewards are equal receive $A=0$. Every response is also scored by the judge of \cref{eq:llr} over its full length.

\paragraph{Win rates.}
The win rates of \cref{fig:mech}a--b use as reference the judge scores of all 5{,}120 sampled responses, including those of groups with zero advantage. The win rate of a subset of responses (the correct or the wrong responses of the groups with a given $k$) is the probability that one of its responses has a higher score than a response drawn from this pool, ties counting half, so 50\% corresponds to a typical response. \Cref{fig:mech}c shows the difference between the correct and the wrong responses of the groups with the same $k$.

\paragraph{Within-group difference.}
Each response is assigned its percentile in the pool. Within every group with non-zero advantages, we take the mean percentile of the correct responses minus that of the wrong responses. The value reported in \cref{sec:mech}, 3.0 percentile points, is the average of this difference over these groups.

\paragraph{First gradient step.}
A LoRA adapter of rank 64 on every linear layer is attached to the 8B original model exactly as at the start of training. Its second matrix is initialized to zero, so the model is identical to the original model and the sampled responses come from the current policy; only the second matrices receive a gradient at this point. The gradient of $\sum_i w_i \log \pi(y_i \mid x_i)$, divided by the total number of response tokens as in training, is accumulated over the 2{,}990 responses of the groups with non-zero advantages, with $w_i = A_i$ for the full signal, $w_i = \max(A_i,0)$ for the positive advantages only, $w_i = \min(A_i,0)$ for the negative advantages only, $w_i = A_{\sigma(i)}$ with a random permutation $\sigma$ within each group for the shuffled control, and $w_i = 1$ for the control in which every response is reinforced equally. Each gradient is turned into the first step of the Adam optimizer, which changes every parameter by $10^{-5}$ in the direction of the sign of its gradient, so that all updates have the same size and differ only in direction. The thinking activation shift $\alpha$ of the updated model is measured on the fixed responses as in \cref{sec:setup}. Every value scales linearly with the step size (a ten-fold step changes it about ten-fold), and rescaling the raw gradient to a common norm instead of taking the Adam step gives the same signs.

\FloatBarrier

\section{Training curves}
\label{app:train}
\begin{figure}[h]
\centering
\includegraphics[width=\linewidth]{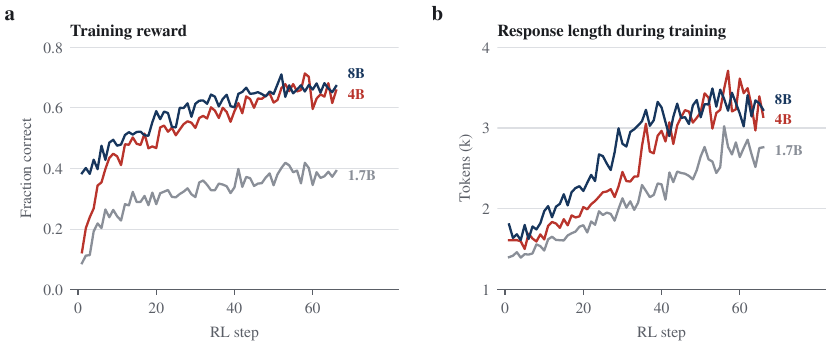}
\caption{Training curves of the three runs: (a) fraction of correct responses on the training batch, (b) mean response length, at each RL step.}
\label{fig:train}
\end{figure}

\Cref{fig:train} shows the training curves of the three runs: the fraction of correct responses on the training batch and the mean response length at each step. The response length grows with the reward at every model size. During training, responses are limited to 8k tokens at 8B and to 16k tokens at 4B and 1.7B. At 8B up to 14\% of the training responses reach this limit late in training; at 4B and 1.7B at most 3\% do.

\FloatBarrier

%% file: tables/amp.tex
\begin{tabular}{@{}cccccccc@{}}
\toprule
\makecell{Update after\\step} & \makecell{Scaling\\factor $c$} & Accuracy & \makecell{Response length\\(tokens)} & Finished & \makecell{Responses with\\\texttt{</think>}} & \makecell{Reflection words\\per response} & $\alpha$ \\
\midrule
15 & 1 & 0.194 & 3.5k & 100\% & 0.0\% & 2.8 & 0.13 \\
15 & 1.77 & 0.294 & 4.5k & 100\% & 0.0\% & 5.0 & 0.26 \\
15 & 2.5 & 0.311 & 6.8k & 100\% & 0.0\% & 12.0 & 0.39 \\
15 & 3.7 & 0.517 & 10.4k & 97\% & 2.8\% & 26.3 & 0.57 \\
15 & 5 & 0.656 & 12.8k & 91\% & 9.4\% & 44.6 & 0.75 \\
15 & 7 & 0.683 & 18.9k & 68\% & 19.4\% & 81.4 & 0.96 \\
15 & 9 & 0.539 & 28.3k & 31\% & 23.3\% & 125.8 & 1.12 \\
15 & 12 & 0.400 & 32.8k & 0\% & 20.6\% & 186.2 & 1.31 \\
67 & 1 & 0.428 & 7.7k & 100\% & 0.0\% & 15.8 & 0.55 \\
67 & 1.5 & 0.572 & 10.5k & 97\% & 6.1\% & 29.7 & 0.79 \\
67 & 2 & 0.694 & 14.2k & 90\% & 19.4\% & 60.0 & 0.98 \\
67 & 3 & 0.722 & 19.2k & 82\% & 50.6\% & 120.5 & 1.25 \\
67 & 4 & 0.611 & 24.7k & 59\% & 53.9\% & 203.2 & 1.42 \\
67 & 5 & 0.417 & 28.4k & 34\% & 40.6\% & 464.5 & 1.55 \\
\bottomrule
\end{tabular}

%% file: tables/lesion_seed2_32k.tex
\begin{tabular}{@{}lcccccc@{}}
\toprule
Condition & Accuracy & \makecell{Response length\\(tokens)} & Finished & \makecell{Reflection words\\per response} & \makecell{Accuracy change\\vs.\ unpatched} & Share of $G$ \\
\midrule
Original model, unpatched & 0.189 & 2.7k & 100\% & 1.6 & -- & -- \\
RL-trained model, unpatched & 0.411 & 7.0k & 100\% & 13.0 & -- & -- \\
\midrule
Original-to-RL patching, think subspace & 0.244 & 4.2k & 100\% & 3.9 & $-0.167$ & 75\% \\
Original-to-RL patching, PCA subspace & 0.361 & 6.2k & 100\% & 10.0 & $-0.050$ & 22\% \\
Original-to-RL patching, random subspace & 0.417 & 6.9k & 100\% & 11.4 & $+0.006$ & $-$3\% \\
Original-to-RL patching, think direction only (1 dimension) & 0.361 & 5.5k & 100\% & 7.8 & $-0.050$ & 22\% \\
RL-to-original patching, think subspace & 0.300 & 4.5k & 100\% & 5.3 & $+0.111$ & 50\% \\
RL-to-original patching, PCA subspace & 0.244 & 3.1k & 100\% & 2.4 & $+0.056$ & 25\% \\
RL-to-original patching, random subspace & 0.211 & 2.8k & 100\% & 1.6 & $+0.022$ & 10\% \\
\bottomrule
\end{tabular}

%% file: tables/steer.tex
\begin{tabular}{@{}ccccc@{}}
\toprule
Strength $a$ & Accuracy & \makecell{Response length\\(tokens)} & Finished & \makecell{Reflection words\\per response} \\
\midrule
0 & 0.189 & 2.8k & 100\% & 1.9 \\
0.25 & 0.200 & 3.6k & 100\% & 4.0 \\
0.5 & 0.289 & 6.2k & 100\% & 12.4 \\
0.75 & 0.467 & 9.2k & 99\% & 32.6 \\
1 & 0.600 & 15.8k & 76\% & 81.2 \\
\bottomrule
\end{tabular}

%% file: tables/surgery.tex
\begin{tabular}{@{}lcc@{}}
\toprule
RL update that is applied & Accuracy & \makecell{Response length\\(tokens)} \\
\midrule
Original model (no update) & 0.189 & 2.6k \\
RL-trained model (full update) & 0.431 & 7.2k \\
Without the part that writes into the think subspace & 0.435 & 7.0k \\
Without the part that writes into a random subspace (same squared norm) & 0.447 & 7.1k \\
Without its top eight singular directions & 0.315 & 5.4k \\
Only in the modules that do not write into the residual stream (q, k, v, gate, up) & 0.343 & 5.4k \\
Only in the modules that write into the residual stream (o, down) & 0.250 & 3.7k \\
\bottomrule
\end{tabular}